\documentclass[letterpaper,twocolumn,10pt]{article}
\usepackage{usenix}
\usepackage{pifont}
\usepackage{booktabs}
\usepackage{amsmath}
\usepackage{graphicx}
\usepackage[ruled,vlined]{algorithm2e}

\newcommand{\LeMo}{\textsc{LeMo}}

\begin{document}

\title{\Large \bf \LeMo{}: Enabling LEss Token Involvement for MOre Context Fine-tuning}

\author{
{\rm Tuowei Wang}\\
Tsinghua University\\
\and
{\rm Xingyu Chen}\\
HUST
\and
{\rm Kun Li}\\
Microsoft Research
\and
{\rm Ting Cao}\\
Microsoft Research
\and
{\rm Ju Ren}\thanks{Corresponding author: Ju Ren (renju@tsinghua.edu.cn)}\\
Tsinghua University
\and
{\rm Yaoxue Zhang}\\
Tsinghua University
} 

\maketitle

\begin{abstract}
The escalating demand for long-context applications has intensified the necessity of extending the LLM context windows. Despite recent fine-tuning approaches successfully expanding context lengths, their high memory footprints, especially for activations, present a critical practical limitation. Current parameter-efficient fine-tuning methods prioritize reducing parameter update overhead over addressing activation memory constraints. Similarly, existing sparsity mechanisms improve computational efficiency but overlook activation memory optimization due to the phenomenon of \textit{Shadowy Activation}.

In this paper, we propose \LeMo{}, the first LLM fine-tuning system that explores and exploits a new token-level sparsity mechanism inherent in long-context scenarios, termed \textit{Contextual Token Sparsity}. \LeMo{} minimizes redundant token involvement by assessing the informativeness of token embeddings while preserving model accuracy. Specifically, \LeMo{} introduces three key techniques: (1) Token Elimination, dynamically identifying and excluding redundant tokens across varying inputs and layers. (2) Pattern Prediction, utilizing well-trained predictors to approximate token sparsity patterns with minimal overhead. (3) Kernel Optimization, employing permutation-free and segment-based strategies to boost system performance. We implement \LeMo{} as an end-to-end fine-tuning system compatible with various LLM architectures and other optimization techniques. Comprehensive evaluations demonstrate that \LeMo{} reduces memory consumption by up to $1.93\times$ and achieves up to $1.36\times$ speedups, outperforming state-of-the-art fine-tuning systems.
\end{abstract}

\section{Introduction}

As the demand for comprehensive document analysis~\cite{document}, extended multi-turn dialogues~\cite{dialogue}, and intricate codebase handling~\cite{code} grows, large language models (LLMs) with larger context windows are becoming integral to AI applications. Despite their utility, LLMs~\cite{chatgpt,mistral,gpt4} are typically pre-trained with fixed context windows, such as the 4K token limit in Llama2~\cite{llama2}. When these models encounter inputs exceeding this limit, their performance deteriorates markedly~\cite{longrope}. The discrepancy between the fixed context window during pre-training and the increasingly extended inputs during inference has emerged as a critical challenge in real-world deployments.

Recent studies~\cite{focused-transformer,long-llama,yarn,scaled-rope} show that the context window of pre-trained LLMs can be extended through fine-tuning on longer sequences. Unfortunately, managing these extended sequences imposes substantial resource challenges, particularly in terms of memory consumption. For instance, Position Interpolation~\cite{position-interpolation} extends Llama models~\cite{llama} from 2K to 32K but requires 128 A100 80GB GPUs, mainly due to memory limitations. Rather than model parameters, the primary memory bottleneck in long-context fine-tuning arises from \textit{activations}~\cite{activation-recompute,activation-recompute-2,lora-fa}, which include intermediate results and gradients that scale proportionally with sequence length.

\begin{table}[t]
    \centering
    \setlength{\tabcolsep}{2.5pt}
    \caption{Memory footprint (GB) comparison across different fine-tuning methods. LoRA and LongLoRA are representative of PEFT and sparsity-based methods, respectively ($S=4K$).}
    \label{tab:memory-consumption}
    \small
    \begin{tabular}{lllll}
        \toprule
        Model             & Llama2-7B     & Llama3-8B     & Mistral-7B    & OPT-6.7B      \\ \midrule
        Naive             & 67.9          & 78.4          & 73.4          & 63.8          \\
        LoRA              & 39.2          & 43.4          & 39.3          & 36.1          \\
        LongLoRA          & 41.3          & 43.9          & 39.3          & 38.1          \\
        \textbf{\LeMo{}} & \textbf{31.3} & \textbf{34.5} & \textbf{31.4} & \textbf{30.0} \\
        \bottomrule
    \end{tabular}
\end{table}

Although various techniques have been proposed to improve the efficiency of fine-tuning, the substantial demands of activation memory remain largely unaddressed. By adapting only a minimal number of parameters, parameter-efficient fine-tuning (PEFT) methods~\cite{lora,adapter,bitfit,prefix-tuning} reduce the memory requirement of parameter updates but leave activation memory unoptimized. Furthermore, recent works~\cite{longlora,long-exposure,infini-attention,seer-attention} incorporate diverse sparse mechanisms to approximate standard dense attention. Despite achieving considerable computational savings, these methods fail to offer additional memory reduction, as highlighted in Table~\ref{tab:memory-consumption}.

\begin{figure}
    \centering
    \includegraphics[width=1.0\linewidth]{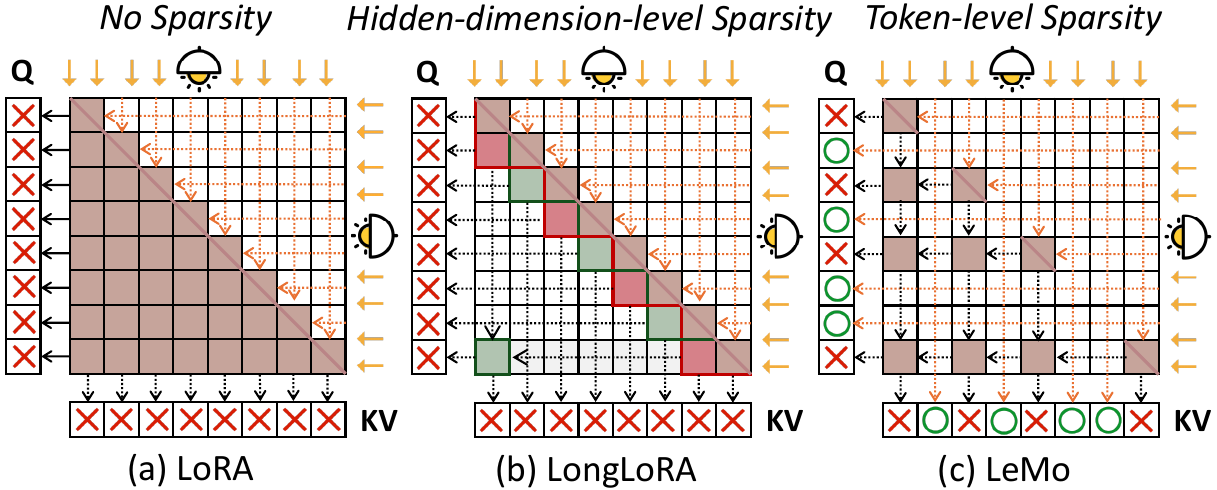}
    \caption{Illustration of shadowy activation. (a) LoRA performs full attention computation without incorporating sparsity. (b) LongLoRA adopts two shifted local attention patterns (colored in red and green) to approximate full attention, targeting sparsity at the hidden-dimension level. (c) \LeMo{} employs token-level sparsity, minimizing token involvement and achieving activation savings compared to other methods.}
    \label{fig:shadowy-activation}
\end{figure}

This limitation arises because existing sparsity mechanisms primarily focus on sparsity within the hidden dimension of individual tokens. As shown in Figure~\ref{fig:shadowy-activation}(b), while each token interacts with a reduced subset of tokens compared to dense attention, the entire token sequence remains involved. Crucially, once a token participates in the computation, regardless of the extent of usage, its activations are retained in memory—a phenomenon we term \textbf{\textit{Shadowy Activation}}. Given the large number of tokens involved in long-context fine-tuning, this observation prompts a critical question: Can we propose a novel LLM fine-tuning scheme with \textit{minimal token involvement} that optimizes \textit{both memory and computational efficiency}?

In this paper, we present \textbf{\LeMo{}}, an efficient system for enhancing long-context fine-tuning of LLMs. The design of \LeMo{} is rooted in an intuitive yet profound observation: natural language exhibits significant redundancy, particularly in long-context scenarios. This redundancy, proved to be ubiquitous in linguistic studies~\cite{text-redundancy,linguistic-redundancy}, has been utilized across multiple LLM-related domains, including data engineering~\cite{data-engineering,metadata}, prompt compression~\cite{selective-context,llmlingua}, and inference optimization~\cite{flexgen,infinigen,h2o,quest}. Most importantly, standard full attention can be effectively approximated by focusing on interactions among a small subset of the most informative tokens in a long text sequence. However, to our knowledge, no research has leveraged this approach to improve fine-tuning efficiency. Therefore, the key insight of \LeMo{} is to \textit{identify and retain only the most informative tokens, enabling long-context fine-tuning with minimized token involvement}.

However, this is not a long-hanging fruit, as this token-level sparsity exhibits unique characteristics in LLM long-context fine-tuning. Specifically, the token embeddings with the highest informativeness vary across different inputs and layers, a mechanism we term \textbf{\textit{Contextual Token Sparsity}}. Harnessing this mechanism effectively and efficiently demands heuristic designs that address both algorithmic and system-level considerations. Specifically, several challenges must be tackled:

\noindent\textbf{Challenge 1: Identification.} While long-context texts inherently exhibit significant redundancy, accurately identifying these redundant tokens is non-trivial. Eliminating informative tokens risks degrading model performance while retaining excessive tokens results in inefficient resource utilization. 

\noindent\textbf{Challenge 2: Detection.} Given the dynamic nature of contextual token sparsity, the optimal sparsity patterns vary across different inputs and layers. This necessitates a selection mechanism capable of adapting to these variations at runtime while incurring minimal computational and memory overhead.

\noindent\textbf{Challenge 3: Performance.} The variability in token selection across layers incurs additional token movements, increasing memory access costs. Besides, naive gradient computation of model loss amplifies activation memory usage, resulting in inefficient memory peaks, especially in long-context scenarios.
 
\LeMo{} consists of three fundamental techniques designed to address these emerging challenges, respectively:

\noindent\textbf{Information-driven Token Elimination.} \LeMo{} evaluates token informativeness by analyzing its interactions with other tokens, selectively excluding less informative ones from computation. Moreover, \LeMo{} exploits sparsity variations across layers by applying layer-specific thresholds, which maximizes resource efficiency while preserving model accuracy.

\noindent\textbf{Context-aware Pattern Prediction.} \LeMo{} utilizes a neural-network-based approach for predicting optimal sparsity patterns that adapt to the context of the current input. Besides, \LeMo{} introduces a technique that elastically transforms the predictor's parameter size, minimizing the memory and computational costs associated with the prediction process.

\noindent\textbf{High-performance Kernel Optimization.} \LeMo{} adapts a permutation-free strategy to eliminate unnecessary global memory data movement during token selection and padding across different layers. Furthermore, \LeMo{} develops a segmented-based gradient computation method, effectively alleviating activation memory peak in long-context scenarios.

We evaluate \LeMo{} across two representative families of LLMs and on three distinct GPU architectures. The results demonstrate that \LeMo{} achieves a memory reduction of $1.93\times$ in end-to-end fine-tuning, compared to the state-of-the-art fine-tuning systems, while maintaining model accuracy.

To the best of our knowledge, \LeMo{} is the first fine-tuning system that exploits the inherent token-level sparsity in long-context scenarios. This innovation allows \LeMo{} to handle long-context sequences while demanding resources for fewer tokens. In summary, we make the following contributions:
\begin{itemize}
    \item We identify a new sparsity mechanism, contextual token sparsity, inherent in long-context fine-tuning, enabling optimizing both memory and computational efficiency.
    \item We develop three key techniques that identify, detect, and perform contextual token sparsity, respectively. Our design provides a heuristic fine-tuning scheme, encompassing both algorithmic and system-level optimizations.
    \item We implement these techniques as an end-to-end fine-tuning system compatible with various LLM architectures, enabling seamless integration with other optimization techniques to further enhance performance.
    \item We conduct comprehensive evaluations of \LeMo{} across diverse LLM families and hardware platforms. The results show that \LeMo{} achieves $1.93\times$ memory savings and $1.36\times$ speedups compared to the state-of-the-art.
\end{itemize}

\section{Background and Motivation}

\subsection{Efficient LLM Fine-tuning}
Fine-tuning, the process of adapting pre-trained LLMs to diverse downstream applications, is pivotal for their effective deployment. Particularly, fine-tuning a pre-trained LLM on longer text sequences is essential for extending its pre-defined context window size, enabling support for long-context scenarios. However, this process is typically resource-intensive, with the inclusion of long text sequences considerably amplifying both computational and memory requirements.

To improve fine-tuning efficiency, one promising direction is parameter-efficient fine-tuning (PEFT) methods~\cite{lora,adapter,bitfit,prefix-tuning}, which adapt pre-trained models by updating only a subset of parameters while maintaining model performance. A representative example is low-rank adaption (LoRA)~\cite{lora}, which freezes the pre-trained model weights and injects smaller, trainable low-rank matrices into each transformer block. By reducing the number of trainable parameters, LoRA markedly alleviates the memory demands of optimizer states, driving its widespread adoption in practical applications.

Another research direction~\cite{seer-attention,informer,longformer,bigbird} focuses on exploiting the inherent sparsity~\cite{sparse-attention} within attention mechanisms by employing various sparsity patterns to the standard dense attention. The central insight is that only a limited subset of interactions is critical, allowing the remainder to be safely disregarded with minimal impact on accuracy. More recent works~\cite{long-exposure,longlora} integrate these two lines of research. Notably, LongLoRA~\cite{longlora} extends LoRA by incorporating a new shifted sparsity pattern, enabling efficient scaling of context length. By substituting global dense attention with two groups of shifted sparse local attention, LongLoRA achieves less training time compared to LoRA. However, subsequent analysis reveals that neither LoRA nor LongLoRA adequately addresses the memory bottleneck in long-context fine-tuning, as activations scale proportionally with sequence length.

\begin{figure}
    \centering
    \includegraphics[width=1.0\linewidth]{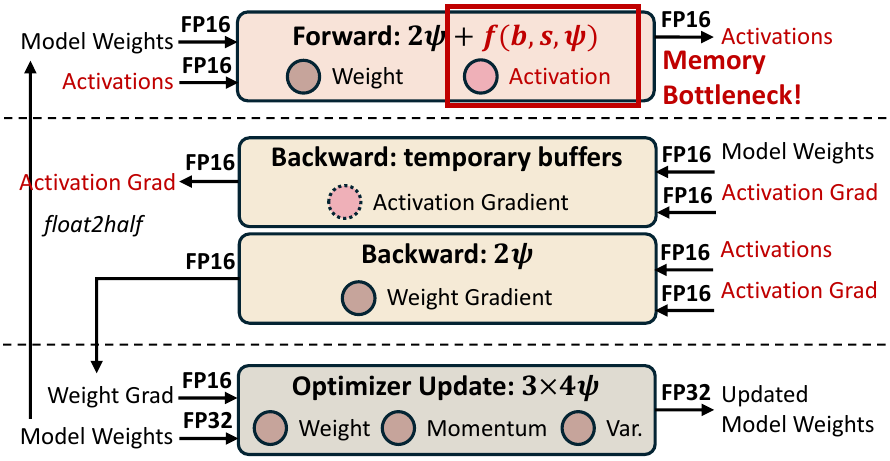}
    \caption{Memory breakdown during LLM mixed-precision fine-tuning. Compared to consistent model states, activations scale with both input batch size and sequence length, becoming the primary bottleneck in long-context scenarios.}
    \label{fig:analysis-vanilla}
\end{figure}

\begin{table}[b]
    \centering
    \setlength{\tabcolsep}{3pt}
    \caption{Activation memory usage (compared to model states) of GPT-3 175B~\cite{gpt3} across different sequence lengths.}
    \label{tab:memory-breakdown}
    \small
    \begin{tabular}{c|lllll}
        \toprule
        Model States        & $s=4K$       & $s=8K$       & $s=16K$      & $s=32K$       & $s=64K$      \\ \midrule
        $16\times175B=2.8T$ & 937G         & 3.42T        & 13.0T        & 50.8T         & 201T         \\
        -                   & 0.34$\times$ & 1.22$\times$ & 4.65$\times$ & 18.14$\times$ & 71.6$\times$ \\
        \bottomrule
    \end{tabular}
\end{table}

\subsection{Analysis: Fine-tuning Memory Breakdown}
\label{sec:memory-breakdown}
Here we provide a detailed analysis of memory consumption during LLM fine-tuning process. Beginning with vanilla fine-tuning, we extend our discussion to include two representative approaches: LoRA, a PEFT method, and LongLoRA, a sparsity-based technique. Our analysis demonstrates that current efficient fine-tuning techniques fall short of addressing memory limitations, particularly in long-context scenarios.

\noindent\textbf{Vanilla.} The memory consumption of vanilla fine-tuning primarily consists of two parts: \textit{model states} and \textit{residual states}, as listed in Figure~\ref{fig:analysis-vanilla}. The first part, \textit{model states}, includes parameters, gradients, and optimizer states. In mixed-precision fine-tuning~\cite{mixed-precision}, parameters and gradients are stored in FP16, while optimizer states are stored in FP32. For modern optimizers like AdamW~\cite{adamW}, the optimizer states include the parameters, momentum, and variance. Given model parameter size as $\psi$, the memory required for all model states is approximately $16\psi$, which remains constant for a fixed model.

The second part of memory consumption comprises activations, temporary buffers, and fragmented memory. Among these, activations, which include the intermediate results and gradients stored during the forward pass, consume a considerable portion of memory. The memory required for activations is not only dependent on the model parameter size ($\psi$) but also scales with the input batch size ($b$) and sequence length ($s$). As detailed in Table~\ref{tab:memory-breakdown}, in long-context scenarios, the memory consumption of activations can easily exceed that of model states, emerging as the dominant memory bottleneck.

\begin{figure}
    \centering
    \includegraphics[width=1.0\linewidth]{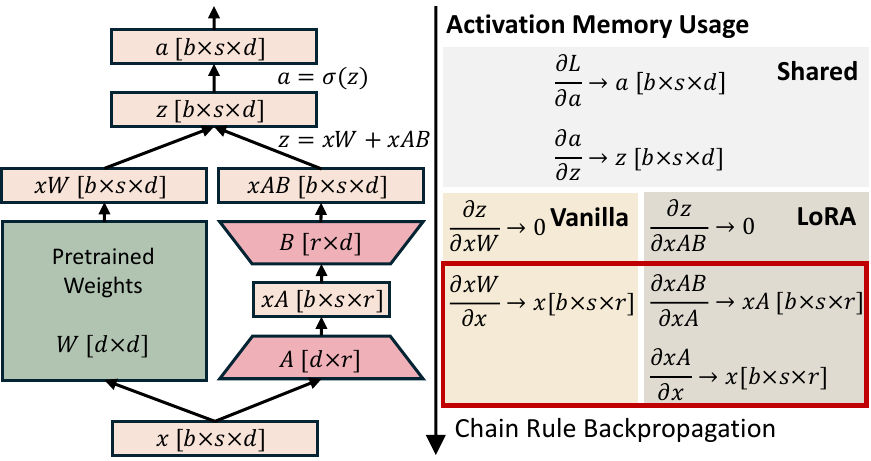}
    \caption{An illustrated example of LoRA for showcasing the activation memory usage of PEFT. Beyond the shared parts, LoRA requires storing more activations than vanilla.}
    \label{fig:analysis-lora}
\end{figure}

\noindent\textbf{LoRA.} Different from the vanilla, the pre-trained model parameters remain frozen in LoRA, and only the injected low-rank matrices are updated during fine-tuning. Consequently, the memory consumption for gradients and optimizer states is significantly reduced. However, this reduction does not extend to activation memory, which emerges as the new memory bottleneck. As illustrated in Figure~\ref{fig:analysis-lora}, LoRA not only fails to alleviate activation memory usage but increases it instead. This is because the trainable low-rank matrices are deeply embedded within the model structure. Gradient computation for these matrices requires a traversal nearly identical to the vanilla, following the chain rule in backpropagation. Similar issues also exist in other PEFT methods.

\noindent\textbf{LongLoRA.} Building upon LoRA, LongLoRA proposes a shifted sparse attention (S$^2$-Attn) mechanism for further efficiency. As depicted in Figure~\ref{fig:shadowy-activation}(b), S$^2$-Attn partitions the input tokens into two groups and performs attention individually in each group. For half of the attention heads, tokens are shifted by half group size to enable information exchange across groups. While LongLoRA achieves computational savings compared to LoRA, it fails to provide additional memory reduction. This limitation arises from the fact that the sparsity mechanism in LongLoRA, as well as other sparsity-based techniques, operates exclusively on the hidden dimensions of token embeddings. While such methods reduce the computational burden associated with individual tokens, they cannot entirely exclude tokens from the computation. However, once a token is involved in the computation, its activations will be stored regardless of its usage. We refer to this phenomenon as \textbf{\textit{Shadowy Activation}}, analogous to how an object casts a shadow when it blocks light. The presence of shadowy activation leaves the activation bottleneck still unresolved.

\begin{figure}
    \centering
    \includegraphics[width=0.92\linewidth]{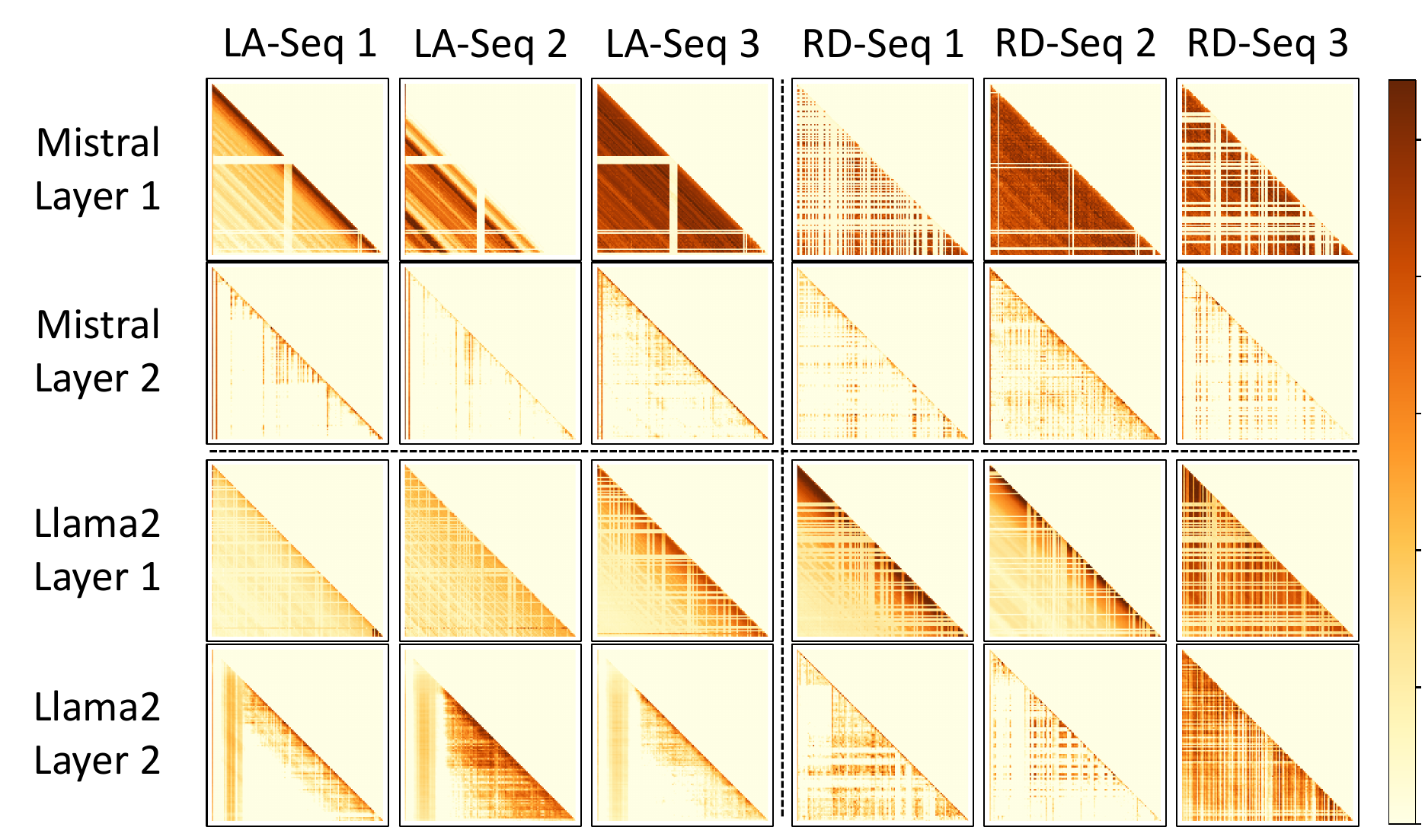}
    \caption{Visualization of attention scores across different models, datasets, layers, and sequences (darker is higher).}
    \label{fig:contextual-token-sparsity}
\end{figure}

\section{Insight: Contextual Token Sparsity}

To address the emerging challenges discussed in \S~\ref{sec:memory-breakdown}, we introduce a fresh perspective that explores and exploits a token-level sparsity mechanism during LLM long-context fine-tuning. Our approach is grounded in the intuitive yet profound observation that natural language is inherently redundant~\cite{text-redundancy,linguistic-redundancy}. Specifically, as highlighted in several studies~\cite{flexgen,infinigen,h2o,quest,retrieval-attention,infllm}, standard full attention can be effectively approximated by focusing on significant interactions among a limited set of query, key, and value values, involving only a subset of tokens in the sequence. Notably, this redundancy becomes even more pronounced in long-context scenarios. As presented in Table~\ref{tab:long-context-sparsity}, the proportion of significant interactions in attention scores decreases as sequence length increases. These insights open up a compelling opportunity to optimize LLM long-context fine-tuning by identifying and retaining only the most informative tokens. Through directly reducing token involvement, shadowy activation constraints are naturally alleviated, enabling activation memory to achieve comparable benefits to computational savings.

\begin{table}[b]
    \centering
    \setlength{\tabcolsep}{2.5pt}
    \caption{Sparsity ratios (the proportion of values below 0.3 of the maximum) of attention scores across sequence lengths.}
    \label{tab:long-context-sparsity}
    \small
    \begin{tabular}{llllllll}
        \toprule
        Seq len. & 4K     & 6K     & 8K     & 10K    & 12K    & 14K    & 16K    \\ \midrule
        Llama2   & 38.6\% & 48.8\% & 48.1\% & 65.3\% & 69.5\% & 62.2\% & 69.6\% \\
        Llama3   & 44.6\% & 46.0\% & 54.7\% & 52.3\% & 57.8\% & 50.2\% & 58.9\% \\
        \bottomrule
    \end{tabular}
\end{table}

We term this novel sparsity mechanism within LLM long-context fine-tuning as \textbf{\textit{Contextual Token Sparsity}}. Through comprehensive evaluations, we identify two key unique characteristics of this mechanism: (1) \textit{Token-wise.} As depicted in Figure~\ref{fig:contextual-token-sparsity}, a grid-like distribution of attention scores is observed across different models and datasets, confirming that token embeddings in the sequence exhibit varying levels of importance. This insight enables the exclusion of less valuable tokens, naturally leading to reductions in both memory usage and computational overhead. (2) \textit{Contextual.} Figure~\ref{fig:contextual-token-sparsity} further demonstrates that the distribution of valuable tokens dynamically shifts based on input texts and varies across model layers, even for a given model and dataset. This dynamic nature underscores the need for a system capable of accurately identifying and efficiently exploiting this sparsity in real-time during runtime. Serving as the cornerstone of our system, contextual token sparsity first introduces token-level sparsity into long-context fine-tuning, enabling LLMs to handle larger context windows while effectively involving fewer tokens.

\begin{figure}
    \centering
    \includegraphics[width=1.0\linewidth]{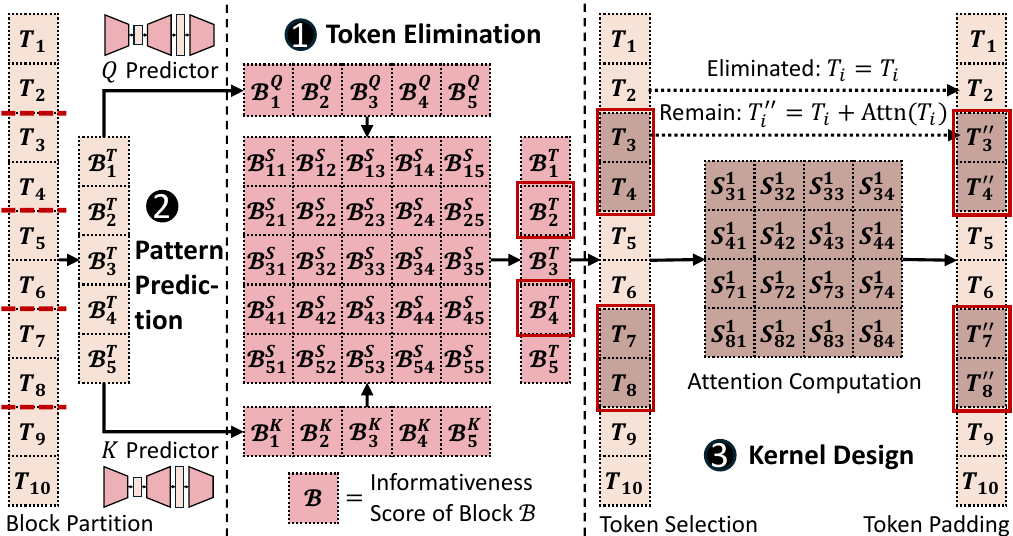}
    \caption{\LeMo{} overview. At each layer, token embeddings are first partitioned into blocks and fed into the pattern predictors (\ding{183}). Using the predicted informativeness scores from these predictors, the token elimination algorithm (\ding{182}) effectively identifies and retains only the most informative tokens for processing. Optimized kernels (\ding{184}) then efficiently perform token selection, computation, residual addition, and padding.}
    \label{fig:overview}
\end{figure}

\section{System Design}

\subsection{Overview of \LeMo{}}
We propose \LeMo{}, an efficient system designed to enhance LLM long-context fine-tuning by systematically exploring and exploiting contextual token sparsity. Figure~\ref{fig:overview} presents an overview of \LeMo{}, which is built upon three key techniques:

\noindent\textbf{Information-driven Token Elimination (\S~\ref{sec:token-elimination}).} To determine whether a token is redundant, we first establish a formal definition for the informativeness of a given token. Building on this definition, \LeMo{} utilizes a score-based algorithm that dynamically identifies and eliminates redundant tokens within the attention block. The algorithm performs in a block-wise manner and is further refined by adopting a layer-specific threshold, which ensures both effectiveness and efficiency. Additionally, we extend this approach to the MLP block, ensuring consistency across various components of the model.

\noindent\textbf{Context-aware Pattern Prediction (\S~\ref{sec:pattern-prediction}).} \LeMo{} employs a neural-network-based approach to predict the token sparsity patterns, bypassing the need for costly full attention score computation. Once adequately trained, these predictors can accurately approximate the token informativeness based on contextual inputs. To minimize the overhead introduced by predictors, \LeMo{} utilizes an elastic size transformation technique, optimizing both memory and computational efficiency.

\noindent\textbf{High-performance Kernel Optimization (\S~\ref{sec:kernel-design}).} \LeMo{} delves into kernel-level optimizations to maximize system performance. To minimize unnecessary global memory movement, \LeMo{} introduces a permutation-free strategy that fuses token selection, token padding, and residual addition directly into the computation pipeline. Moreover, \LeMo{} incorporates a segment-based method to alleviate activation memory peaks, enabling efficient gradient computation for long sequences. By tightly coupling these kernel-level optimizations with algorithmic designs, \LeMo{} fully exploits contextual token sparsity for enhancing LLM long-context fine-tuning.

\begin{figure}
    \centering
    \includegraphics[width=1.0\linewidth]{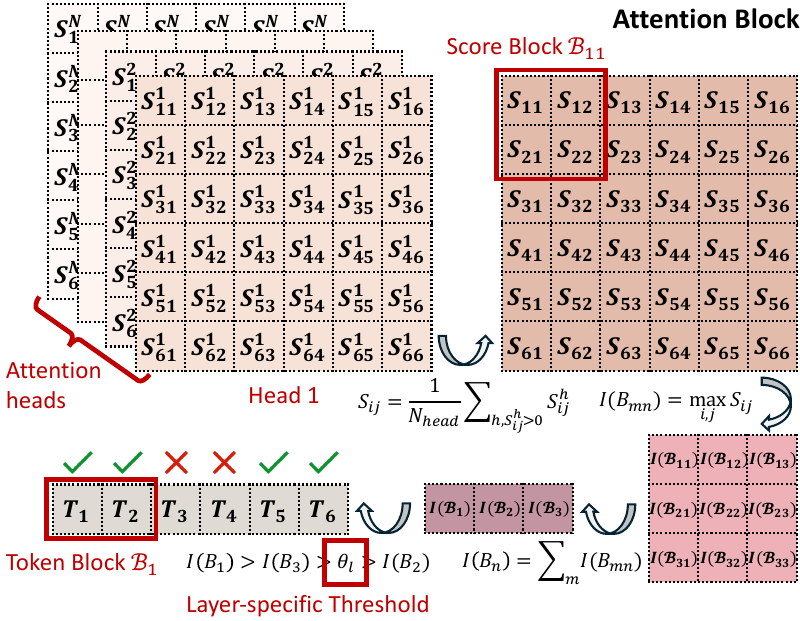}
    \caption{Token elimination algorithm. Attention scores are first aggregated across different heads and partitioned into multiple score blocks. The maximum value within each score block is defined as its informativeness score, which is then aggregated along the column. These resulting scores are compared against a layer-specific threshold to determine whether the corresponding tokens should be retained.}
    \label{fig:algorithm}
\end{figure}

\subsection{Information-driven Token Elimination}
\label{sec:token-elimination}
To fully exploit contextual token sparsity, it is essential to accurately identify the redundant tokens across different inputs and layers. Discarding informative tokens may impact model accuracy while retaining excessive tokens leads to resource inefficiency. To tackle these challenges, \LeMo{} proposes an information-driven algorithm that dynamically identifies and eliminates redundant tokens while preserving accuracy.

\noindent\textbf{Token Informativeness.} We begin by defining a token's informativeness based on its interactions with other tokens within the embedding space. In the attention mechanism, the attention score $S_{\text{attn}}$ is commonly used to quantify the interaction between tokens~\cite{powerbert,infllm,retrieval-attention}. Specifically, the attention score term $S_{ij}$, calculated as $Q_{i}K_{j}$, represents the interaction between token $i$ and token $j$. Inspired by this, we define the informativeness of a token $I(T)$ by considering its interaction with all other tokens in the long-context sequence:
\begin{equation}
    I(T_{j}) = \sum_{i \neq j}S_{ij} = \sum_{i \neq j}{Q_{i}K_{j}}
\end{equation}
where the sum aggregates the attention scores across all tokens $i$ in the sequence, excluding the token $j$ itself.

\noindent\textbf{Block-wise Elimination.} Building on the concept of token informativeness, the next step involves eliminating redundant tokens based on their informativeness scores. To optimize alignment with hardware characteristics, \LeMo{} performs token elimination in a block-wise manner. As shown in Figure~\ref{fig:algorithm}, \LeMo{} partitions the attention scores along the token dimension into multiple \textit{score blocks $\mathcal{B}^{S}$}. Following aggregation across attention heads, the maximum value within each block is selected as the informativeness score $I(\mathcal{B}^{S})$. Notably, during aggregation, \LeMo{} sums only the positive attention scores. Since attention scores undergo a softmax operation, negative values have a negligible impact on the final result but may offset the influence of positive values if included. By excluding negative scores, the aggregation process preserves the integrity of informativeness and ensures robust token elimination. The entire process can be formalized as:
\begin{equation}
    I(\mathcal{B}_{mn}^{S}) = \max_{S_{ij} \in \mathcal{B}_{mn}^{S}}S_{ij} = \max_{S_{ij} \in \mathcal{B}_{mn}^{S}}\frac{\sum_{h,S_{ij}^{h} > 0} S_{ij}^{h}}{N_\text{head}}
\end{equation}
where $S_{ij}^{h}$ represents the attention score between token $i$ and token $j$ in attention head $h$, and $N_\text{head}$ is the total number of attention heads, serving as the scaling factor.

\begin{algorithm}[b]
    \caption{Layer-Specific Threshold Optimization}
    \label{alg:layer-specific-threshold}
    \small
    \SetKwInOut{Input}{Input}
    \SetKwInOut{Output}{Output}
    \Input{Model layers $\mathcal{L} = \{L_1, L_2, \dots, L_n\}$}
    \Output{Layer thresholds $\mathcal{T} = \{T_1, T_2, \dots, T_n\}$}
    \BlankLine
    \tcp{Step 1: Threshold Initialization}
    \ForEach{layer $L_i \in \mathcal{L}$}{
        \tcp{Average scores across token blocks}
        $T_i \gets \text{avg}\big(I(\mathcal{B}^{T}) \;\forall\; \mathcal{B}^{T} \in L_i\big)$\;
    }
    \tcp{Step 2: Threshold Fine-Tuning}
    \ForEach{layer $L_i \in \mathcal{L}$}{
        \tcp{Compute gradient with finite changes}
        $G_i \gets \displaystyle \big(\mathrm{acc}(T_i + \epsilon) - \mathrm{acc}(T_i - \epsilon)\big)/2\epsilon$\;
        \tcp{Update threshold based on gradient}
        $T_i \gets T_i + \eta \cdot G_i$\;
    }
    \Return $\mathcal{T}$
\end{algorithm}

\begin{figure}
    \centering
    \includegraphics[width=1.0\linewidth]{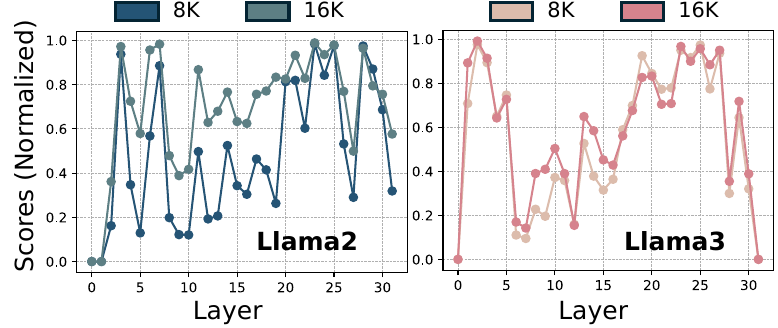}
    \caption{The average informativeness scores (normalized) of tokens blocks across different layers.}
    \label{fig:layer-specific-sparsity}
\end{figure}

\noindent\textbf{Layer-specific Threshold.}
The informativeness scores of \textit{token blocks $\mathcal{B}^{T}$} are computed by aggregating the corresponding score blocks, i.e. $\mathcal{B}_{n}^{T}=\sum_{m}\mathcal{B}_{mn}^{S}$. These scores are then compared against a threshold to determine whether the tokens within the block should be eliminated. Particularly, \LeMo{} further refines the token elimination algorithm by adopting a layer-specific threshold. The key insight is that different layers within LLMs exhibit varying sparsity patterns. As demonstrated in Figure~\ref{fig:layer-specific-sparsity}, the average informativeness scores of token blocks vary greatly across different model layers, indicating that a universal threshold applied across all layers is suboptimal. Algorithm~\ref{alg:layer-specific-threshold} outlines \LeMo{}'s approach, which initializes a default threshold for all layers based on score profiling and then fine-tunes these values to align with the unique sparsity characteristics of each layer.

\noindent\textbf{Extend to MLP Block.} Additionally, we extend token elimination to the MLP block. Analogous to attention scores, \LeMo{} utilizes intermediate activations within the MLP block to evaluate the informativeness of each token. This extension can be viewed as a variant of the widely studied neuron sparsity within MLP blocks~\cite{activation-sparsity,lazy-neuron,relu-strikes-back}, ensuring compatibility with various MLP block structures: for ReLU-based structure~\cite{relu}, the activations are the outputs of the ReLU layer, while for SiLU-based structure~\cite{silu}, the activations correspond to the element-wise multiplication of the gate projection (after SiLU) and the up projection. These techniques enable \LeMo{} to adapt token elimination seamlessly across diverse model components and configurations.

\subsection{Context-aware Pattern Prediction}
\label{sec:pattern-prediction}
Although the exact sparsity patterns can be directly derived from the full attention scores, computing and storing these scores is prohibitively expensive, with complexity scaling quadratically with the sequence length. Furthermore, due to the dynamic nature of contextual token sparsity, the optimal sparsity patterns can only be determined at runtime, varying across different inputs and layers. To address these challenges, \LeMo{} employs a set of lightweight neural networks as predictors. By taking contextual embeddings as inputs, these predictors infer sparsity patterns accurately and efficiently.

\begin{figure}
    \centering
    \includegraphics[width=1.0\linewidth]{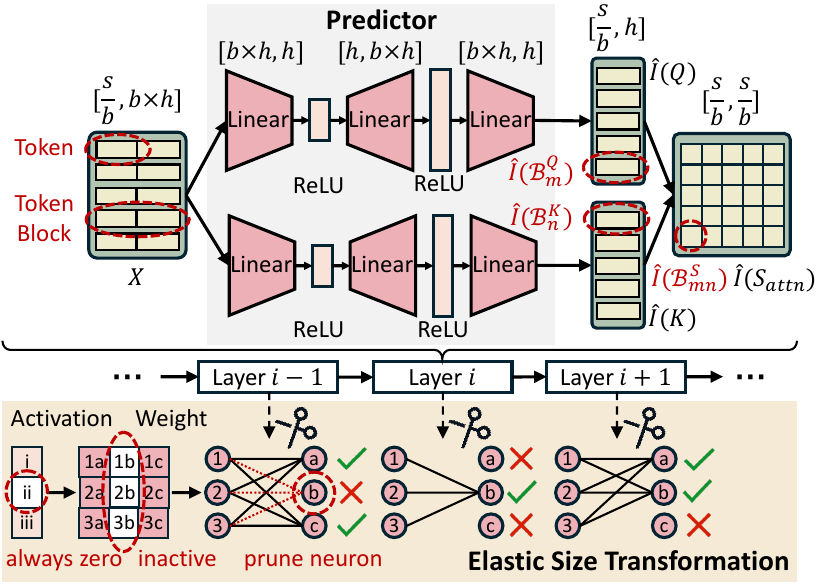}
    \caption{Pattern prediction process. Each layer is equipped with two predictors to approximate $Q$ and $K$, respectively. Taking token embeddings as input (organized in token blocks), each predictor outputs the informative score, $\hat{I}(B_{m}^{Q})$ and $\hat{I}(B_{n}^{K})$, for each token blocks. These scores are then multiplied to compute the informative scores, $\hat{I}(B_{mn}^{S})$, for blocked attention scores. Besides, elastic size transformation is employed to independently minimize the predictor size for each layer.}
    \label{fig:predictor}
\end{figure}

\noindent\textbf{Neural-network-based Predictor.} As illustrated in Figure~\ref{fig:predictor}, \LeMo{} deploys a pair of predictors in each layer to approximate the informativeness scores of queries $Q$ and keys $K$, respectively. Each predictor consists of three trainable low-rank matrices, with ReLU activation function applied between successive matrices. The inputs to the predictors are token embeddings $X$, which contain contextual information and are organized into blocks to align with the block-wise elimination. By extracting the representative embedding from each block, the predictors output the approximate informativeness scores, $\hat{I}(Q)$ and $\hat{I}(K)$. These scores are then multiplied to approximate the informativeness of attention scores $\hat{I}(S_{\text{attn}})$:
\begin{equation}
        \hat{I}(S_{\text{attn}}) = \hat{I}(Q)\hat{I}(K)^{T},\ \hat{I}(\mathcal{B}_{mn}^{S})=\hat{I}(\mathcal{B}_{m}^{Q})\hat{I}(\mathcal{B}_{n}^{K})^{T}
\end{equation}
When $Q$ and $K$ predictors are well trained, $\hat{I}(S_{\text{attn}})$ can provide
a close estimation of accurate informativeness scores $I(S_{\text{attn}})$.

With a limited training dataset, these predictors can quickly converge and perform well in prediction, as evidenced by prior studies~\cite{dejavu,powerinfer,long-exposure}. Particularly, the predictor in \LeMo{} first processes each token individually and then aggregates these individual predictions into a unified outcome. This strategy restricts the predictor’s size to the dimension of a single token block rather than the full long-context sequence, thereby streamlining the design and mitigating prediction overhead.

\noindent\textbf{Elastic Size Transformation.} To further minimize the predictor's size, \LeMo{} utilizes an elastic size transformation technique that dynamically prunes neurons in the predictors. This design leverages the properties of the ReLU activation function, which introduces a significant number of zero elements into the intermediate activation of the predictors. When an activation element is zero, its corresponding neurons (i.e., rows or columns of the model weights) become inactive and can be safely disregarded. Building on this insight, \LeMo{} tracks the zero frequency of intermediate activation elements during training and periodically prunes neurons associated with the highest zero frequencies. Without relying on any prior assumptions, elastic size transformation adaptively determines the optimal size for each predictor, simultaneously reducing both computational and memory overhead.

\noindent\textbf{Comprehensive Overhead Analysis.} We conclude with an analysis of the overhead introduced by predictors during both training and inference. In \textit{offline training}, the primary bottleneck lies in obtaining the informativeness of attention scores $I(S_{\text{attn}})$. Thanks to the block-wise manner, we seamlessly integrate our custom training kernel into the state-of-the-art FlashAttention~\cite{flash-attention,flash-attention2}. This integration eliminates the need for explicit computation and storage of the full attention scores. Instead, $I(S_{\text{attn}})$ is derived online, leading to a memory complexity that grows linearly with sequence length.

In \textit{online inference}, given sequence length $s$, head dimension $h$, and block size $b$, the computational overhead consists of two parts: (1) Prediction of $I(Q)$ or $I(K)$ with a complexity of $O(sh^2)$; (2) Prediction of $I(S_{\text{attn}})$ with a complexity of $O(s^2/b^2)$. In long-context scenarios, the second part becomes the dominant factor. However, this overhead can be effectively mitigated by increasing the block size $b$. On the memory side, the primary overhead arises from the linear weights within predictors, with a complexity of $O(bh^2)$. Importantly, this complexity remains constant relative to the model configurations. Finally, thanks to the elastic size transformation, both computational and memory complexities are reduced by a sparsity factor, leading to an average reduction of 50\%. 

\subsection{High-performance Kernel Optimization}
\label{sec:kernel-design}
Focusing on token-level sparsity, \LeMo{} introduces minimal modifications to the original fine-tuning dynamics, enabling seamless reuse of existing optimized computational flows. However, two key challenges hidden in \LeMo{} undermine its performance. First, the variability in sparsity patterns across layers necessitates iterative token selection and padding, leading to considerable costly global memory movement. Second, the extensive vocabulary size of LLMs necessitates substantial activation memory to compute the output loss gradient for each token, particularly in long-context scenarios. \LeMo{} incorporates several hardware-efficient techniques at the kernel level, effectively mitigating these bottlenecks.

\begin{figure}
    \centering
    \includegraphics[width=1.0\linewidth]{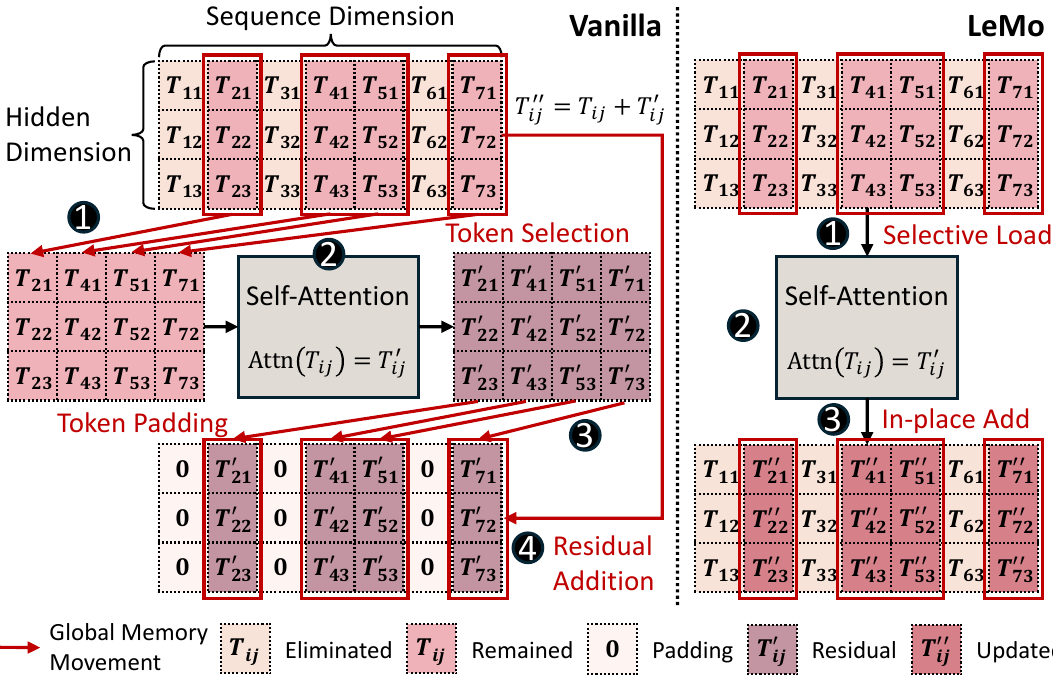}
    \caption{Comparison between naive token movement and \LeMo{}. Highlighted by the red lines, the naive kernel incurs substantial global memory movement costs. \LeMo{} develops a permutation-free strategy through kernel fusion.}
    \label{fig:kernel-movement}
\end{figure}

\noindent\textbf{Permutation-free Token Movement.} The dynamic nature of contextual token sparsity leads to varying sparsity patterns across different layers, involving distinct subsets of tokens. As illustrated in Figure~\ref{fig:kernel-movement}, at each layer, a set of less informative tokens are eliminated and the remained tokens are re-permuted to serve as the attention block inputs. Then, the attention outputs are padded with zeros to maintain dimensional consistency and are finally residually added to the original inputs. The processes of token selection, token padding, and residual addition involve extensive data movement in global memory, which incurs high memory access latency.

\LeMo{} develops a permutation-free strategy that fuses all unnecessary permutation operations with the attention computations. Instead of materializing re-permuted tokens, \LeMo{} directly loads the selected tokens from original inputs. Furthermore, \LeMo{} performs an in-place addition of attention outputs to the original inputs, simultaneously completing token padding and residual addition in a single step. This streamlined approach eliminates expensive global memory movement, significantly enhancing system performance.

\begin{figure}
    \centering
    \includegraphics[width=1.0\linewidth]{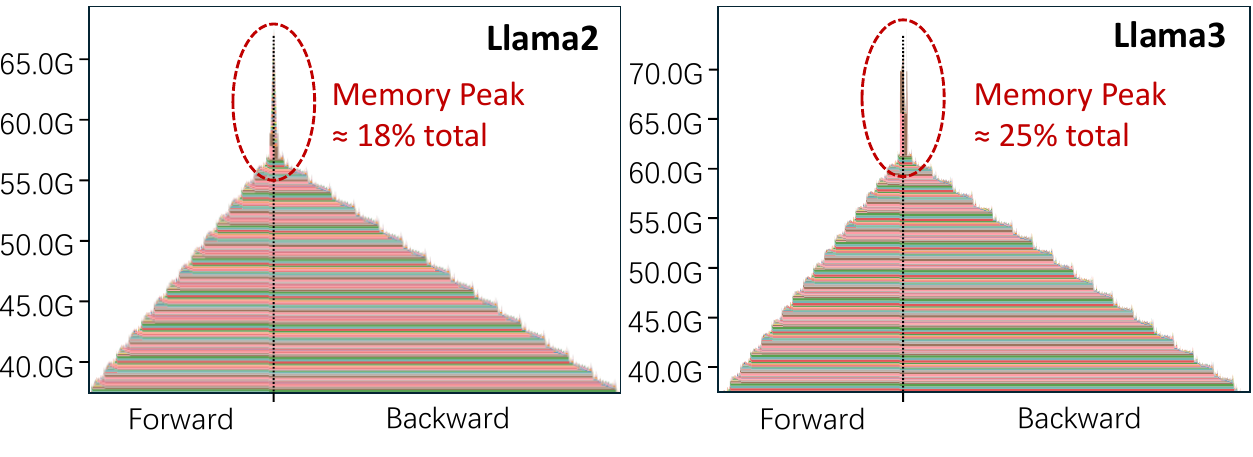}
    \caption{Memory peak during loss gradient computation, exacerbated by the large vocabulary size and long context.}
    \label{fig:kernel-cutting}
\end{figure}

\noindent\textbf{Segment-based Peak Cutting.} LLMs are typically auto-regressive, predicting the probability distribution of the next token given all preceding tokens. During fine-tuning, for each token in the input sequence, the model generates a probability distribution over the next token and computes the losses between predictions and ground truth. For LLMs with a large vocabulary size and long context window, this process imposes a significant surge in activation memory usage, as shown in Figure~\ref{fig:kernel-cutting}. While these activations are transient, the resulting memory peak elevates the upper bound of LLM fine-tuning, imposing stricter demands on GPU memory resources.

To address this issue, \LeMo{} adopts a segment-based peak-cutting strategy that partitions the token sequence into smaller, manageable segments during final loss computation. Instead of performing a forward pass over the entire sequence and retaining all intermediate activations, \LeMo{} processes each segment independently and later aggregates their gradients. Activations for each segment are discarded immediately after the corresponding gradient computation is completed. Consequently, the activation memory peak is reduced to $1/N$ when the sequence is divided into $N$ segments. This approach greatly alleviates memory pressure, enabling efficient long-context fine-tuning within constrained GPU memory limits.

\begin{figure}
    \centering
    \includegraphics[width=1.0\linewidth]{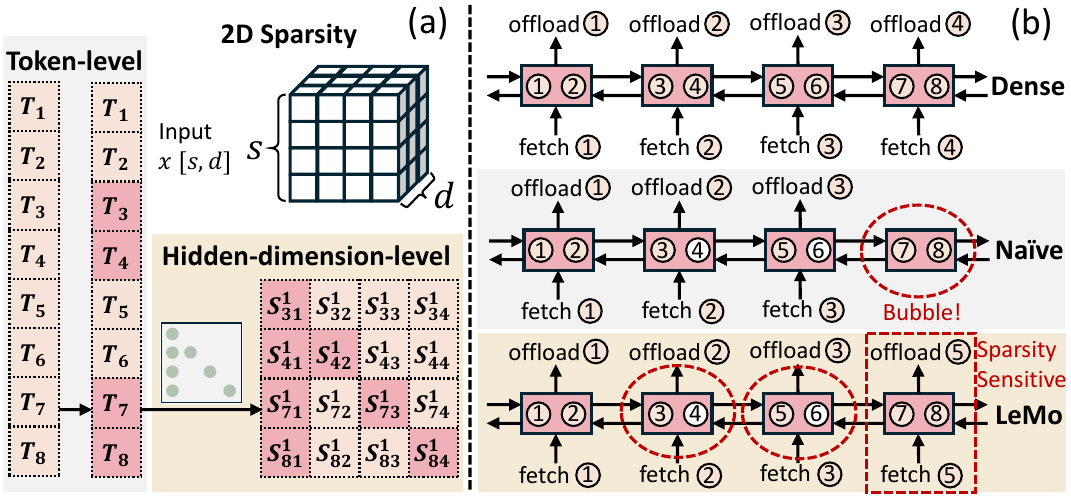}
    \caption{Two available extensions of \LeMo{}: (a) Two-dimensional Sparsity and (b) Sparsity-sensitive Offload.}
    \label{fig:extension}
\end{figure}

\section{Implementation and Extension}

We implement \LeMo{} with over 3000 lines of Python and C++ code. Given the minimal changes to the original fine-tuning dynamics, \LeMo{} is compatible with a wide range of LLM architectures without requiring any code changes. Besides, \LeMo{} can seamlessly integrate with other techniques:

\noindent\textbf{Extension 1: Two-dimensional Sparsity.} As illustrated in Figure~\ref{fig:extension}(a), after applying token-level sparsity in \LeMo{}, the remained tokens can further benefit from existing hidden-dimensional sparsity techniques. This natural combination of sparsity mechanisms across two dimensions, which we term \textit{2D-Sparsity}, provides more granular control over the model’s resource allocation, leading to a significant reduction in both activation memory and computational costs.

\noindent\textbf{Extension 2: Sparsity-sensitive Offload.} \LeMo{} enhances existing offload-based techniques~\cite{autotm,swapadvisor,capuchin,l2l} by incorporating contextual token sparsity into the optimization process. As depicted in Figure~\ref{fig:extension}(b), we develop a sparsity-sensitive offloading strategy that adapts to the varying sparsity ratios across different layers. This approach enables the seamless transfer of larger data volumes between the CPU and GPU, effectively alleviating GPU memory constraints.

\section{Evaluation}

\subsection{Experimental Setup}
\noindent\textbf{Hardware.} We conduct experiments on three representative platforms, as listed in Table~\ref{tab:exp-setup-hardware}, covering both data-center workstations and desktop professional GPUs. Memory measurements are primarily evaluated on Platform A, as it offers the largest GPU memory capacity, with results being largely insensitive to GPU arithmetic performance. For speedup evaluations, we adhere to common practices by employing mixed-precision techniques~\cite{mixed-precision}, utilizing both BF16 and FP32.

\begin{table}[t]
    \centering
    \setlength{\tabcolsep}{2.2pt}
    \caption{Hardware platform configurations.}
    \label{tab:exp-setup-hardware}
    \small
    \begin{tabular}{lllcc}
        \toprule
        Platform   & GPUs           & Memory & FP32 TFLOPS & BF16 TFLOPS \\ \midrule
        Platform A & $1\times$ A800 & 80GB   & 19.5        & 312  \\
        Platform B & $1\times$ A40  & 48GB   & 37.4        & 150  \\
        Platform C & $4\times$ 4090 & 24GB   & 82.6        & 82.6 \\
        \bottomrule
    \end{tabular}
\end{table}

\begin{table}[t]
    \centering
    \caption{Model configurations.}
    \label{tab:exp-setup-models}
    \small
    \begin{tabular}{llll}
        \toprule
        Model                  & \# Params.          & Def Len. & Seq Len. \\ \midrule
        OPT~\cite{opt}         & 350M/1.3B/2.7B/6.7B & 2K       & 2K-64K   \\
        Llama2~\cite{llama2}   & 7B                  & 4K       & 4K-64K   \\
        Llama3~\cite{llama3}   & 8B                  & 8K       & 4K-64K    \\
        \bottomrule
    \end{tabular}
\end{table}

\noindent\textbf{Models.} The models used for evaluation are detailed in Table~\ref{tab:exp-setup-models}. We choose models from two of the most popular LLM families: OPT and Llama. These models vary in architecture, parameter size, and default context window size. Evaluations on these models provide a robust demonstration of the scalability and versatility of \LeMo{}'s optimizations.

\noindent\textbf{Dataset.} We use the RedPajama~\cite{red-pajama} dataset for long-context fine-tuning, following the setup from LongLoRA. We evaluate the perplexity (PPL) of fine-tuned models on the book corpus dataset PG19~\cite{pg19} and the cleaned Arxiv Math proof-pile dataset~\cite{proof-pile} to assess long-context modeling performance. Besides, we evaluate our method on LongBench~\cite{longbench} benchmark, following an instruction-tuning on dataset LongAlign-10k~\cite{longalign}. These tasks cover multiple critical long-text application areas, ensuring a comprehensive evaluation of our approach.

\noindent\textbf{Baselines.} We compare \LeMo{} with two state-of-the-art fine-tuning methods, LoRA~\cite{lora} and LongLoRA~\cite{longlora}. These methods represent the two dominant optimization directions: parameter-efficient fine-tuning and sparsity-based fine-tuning, respectively. For speedup analysis, we primarily compare with LoRA, leaving LongLoRA for reference. This distinction is because the two methods focus on orthogonal sparsity dimensions, making a direct and fair comparison challenging.

\noindent\textbf{Metrics.} Memory evaluations are conducted at the time step immediately following the forward pass, which generally corresponds to the memory peak. For execution time evaluation, we measure the time per fine-tuning step. To emphasize the effectiveness of \LeMo{}, activation recomputation and offloading techniques are excluded unless explicitly stated. All metrics are averaged over 10 repeated trials to ensure reliability.

\subsection{End-to-End Performance}
\begin{figure}
    \centering
    \includegraphics[width=1.0\linewidth]{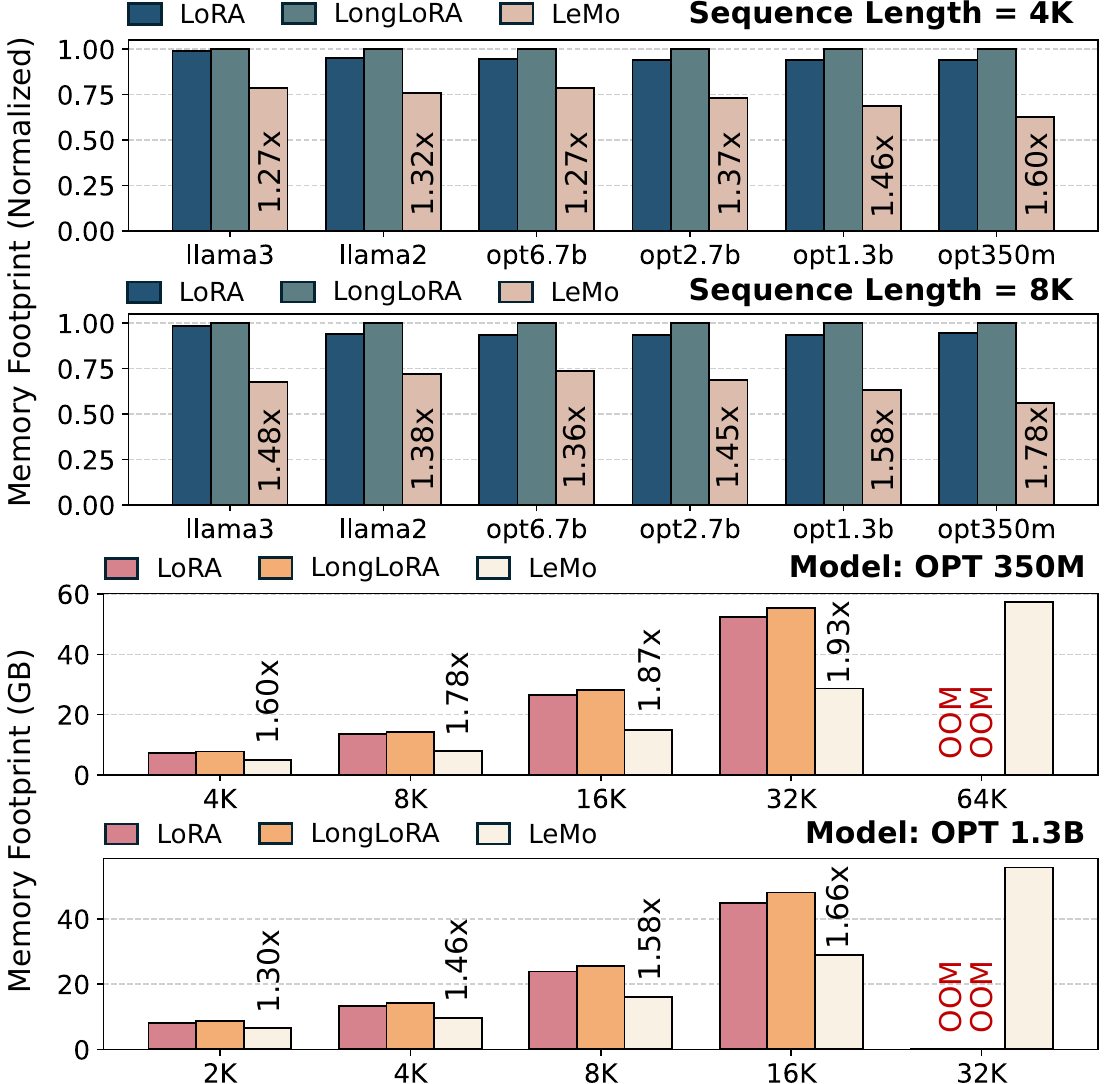}
    \caption{Memory footprints comparison on A800.}
    \label{fig:exp-end2end-memory}
\end{figure}

\noindent\textbf{Memory Footprint.} We evaluate the memory efficiency of \LeMo{} across various models and sequence lengths, as shown in Figure~\ref{fig:exp-end2end-memory}. The results reveal that \LeMo{} achieves average memory savings of 38.2\% and 50.5\% compared to LoRA across six different models, with sequence lengths of 4K and 8K, respectively. Similar benefits are observed in comparison to LongLoRA, as the presence of shadowy activations renders its sparsity mechanism ineffective in benefiting memory usage (even slightly increased). Furthermore, the results reveal that for a fixed model, \LeMo{}'s memory efficiency improves as sequence length increases. This aligns with the observation that longer text sequences typically exhibit greater redundancy. The enhanced efficiency of \LeMo{} largely extends the fine-tuning sequence length achievable under GPU memory constraints. Without activation recomputation and offloading, both LoRA and LongLoRA are limited to a sequence length of 16K (32K) when fine-tuning OPT 1.3B (350M). Instead, \LeMo{} doubles this capacity, supporting sequence lengths of up to 32K (64K) on a single A800 GPU.

\begin{figure}
    \centering
    \includegraphics[width=1.0\linewidth]{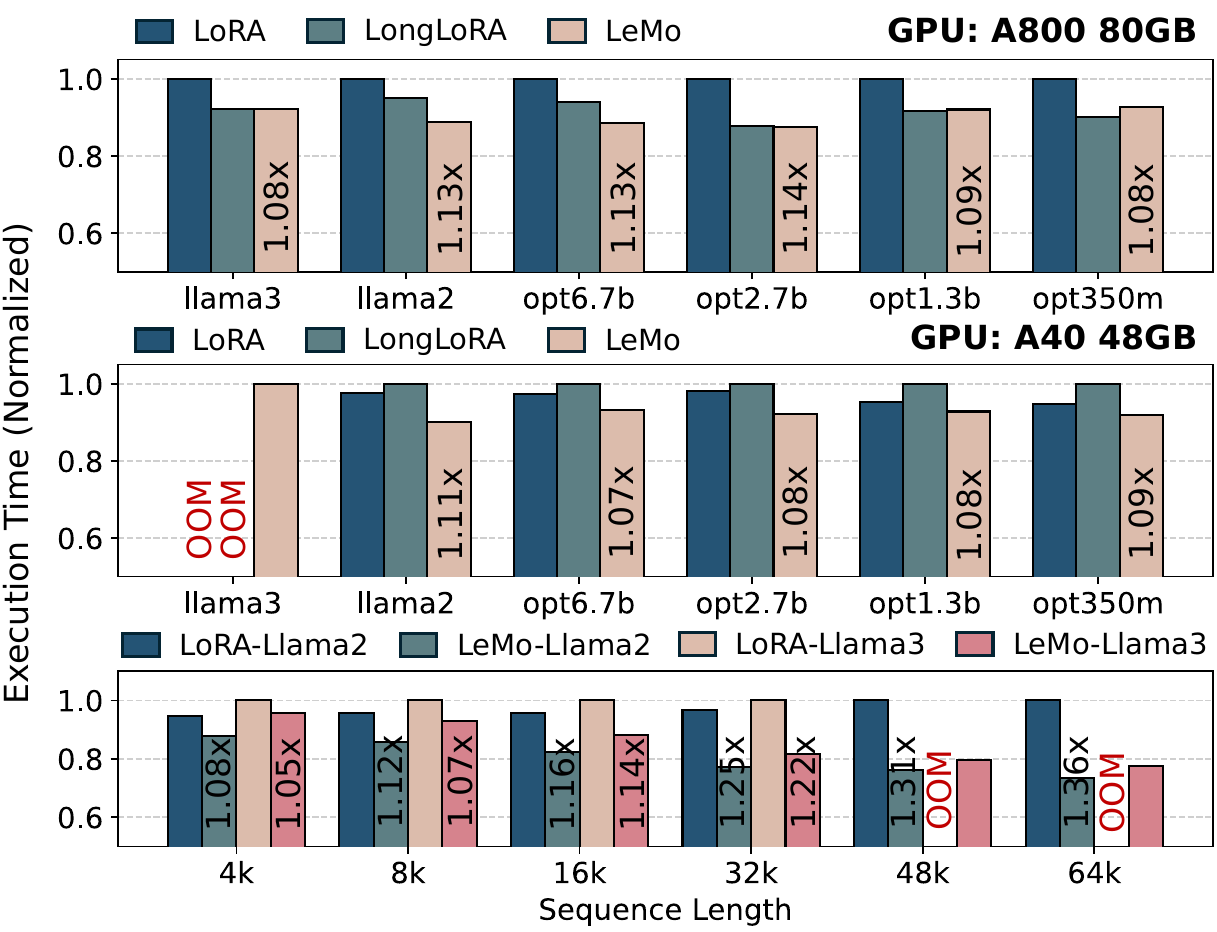}
    \caption{End-to-end speedup of \LeMo{} on A800 and A40.}
    \label{fig:exp-end2end-time}
\end{figure}

\noindent\textbf{Execution Time.} The minimized token involvement in \LeMo{} also brings computational savings during long-context fine-tuning. Figure~\ref{fig:exp-end2end-time} presents the execution time and corresponding speedups of \LeMo{} during fine-tuning different models at a sequence length of 4K. The results show that \LeMo{} achieves computational efficiency comparable to LongLoRA, achieving an average speedup over LoRA of 10.8\% and 8.6\% on two platforms, respectively. We also observe that LongLoRA may perform slower than LoRA in some cases. This is primarily due to the difficulty in fully utilizing hardware computational capacity for sparsity operations when the sequence length is insufficient. In contrast, \LeMo{} introduces minimal modifications to the original computational flow, allowing it to effectively translate computational savings into practical speedups. Further evaluations of \LeMo{} on longer sequence lengths (with recomputation) reveal additional performance gains, achieving up to $1.36\times$ speedups.

\begin{table*}[t]
    \centering
    \setlength{\tabcolsep}{2pt}
    \caption{Comparative analysis of model accuracy on the LongBench benchmark (higher is better).}
    \label{tab:exp-end2end-accuracy-benchmark}
    \small
    \begin{tabular}{llllllllll}
        \toprule
        Tasks  & mfqa\_zh       & mfqa\_en       & gov\_report    & triviaqa       & vcsum          & qmsum          & musique       & 2wikimqa       & repobench      \\ \midrule
        Origin & 23.45          & 23.22          & \textbf{27.44} & \textbf{84.60} & \textbf{13.30} & \textbf{22.64} & 4.63          & 9.01           & \textbf{52.00} \\
        Ours   & \textbf{23.53} & \textbf{24.74} & 25.92          & 82.59          & 13.02          & 20.33          & \textbf{5.73} & \textbf{10.14} & 48.32          \\ \midrule
        Tasks  & qasper         & hotpotqa       & multi\_news    & pr\_zh         & pr\_en         & trec           & lsht          & dureader       & lcc            \\ \midrule
        Origin & 15.94          & 9.40           & \textbf{24.43} & \textbf{10.0}  & 20.0           & \textbf{68.0}  & 21.0          & \textbf{23.69} & \textbf{71.28} \\
        Ours   & \textbf{17.68} & \textbf{9.55}  & 22.53          & 8.00           & \textbf{22.0}  & \textbf{68.0}  & \textbf{25.0} & 21.37          & 70.32          \\
        \bottomrule
    \end{tabular}
\end{table*}

\begin{table}[t]
    \centering
    \caption{Perplexity of PG19 (PG) and Proof-Pile (PP) datasets.}
    \label{tab:exp-end2end-accuracy-ppl}
    \small
    \begin{tabular}{llllll}
        \toprule
        Seq len.    & 8K   & 10K  & 12K  & 14K  & 16k  \\ \midrule
        PG-Origin   & 6.95 & 6.92 & 6.91 & 6.90 & 6.87 \\
        PG-Ours     & 7.11 & 7.12 & 7.08 & 7.13 & 7.08 \\ \midrule
        PP-Origin   & 2.68 & 2.64 & 2.59 & 2.59 & 2.57 \\
        PP-Ours     & 2.79 & 2.77 & 2.72 & 2.72 & 2.70 \\
        \bottomrule
    \end{tabular}
\end{table}

\noindent\textbf{Accuracy Evaluation.} We test the impact of \LeMo{} on model accuracy by comparing it with the original LoRA. First, we measure test perplexity of fine-tuned Llama2 7B on two representative long-context datasets, PG19 and Proof-Pile. As shown in Table~\ref{tab:exp-end2end-accuracy-ppl}, \LeMo{} incurs only a minimal increase in perplexity scores compared to the original LoRA, across varying sequence lengths. Additionally, we evaluate \LeMo{} on LongBench benchmark, which contains tasks from multiple key long-text application areas. Table~\ref{tab:exp-end2end-accuracy-benchmark} demonstrates that \LeMo{} achieves accuracy comparable to the original LoRA. These evaluations collectively confirm that the inherent redundancy in long-context sequences can be effectively exploited for performance efficiency without compromising accuracy.

\subsection{Ablation Study}
\begin{figure}
    \centering
    \includegraphics[width=1.0\linewidth]{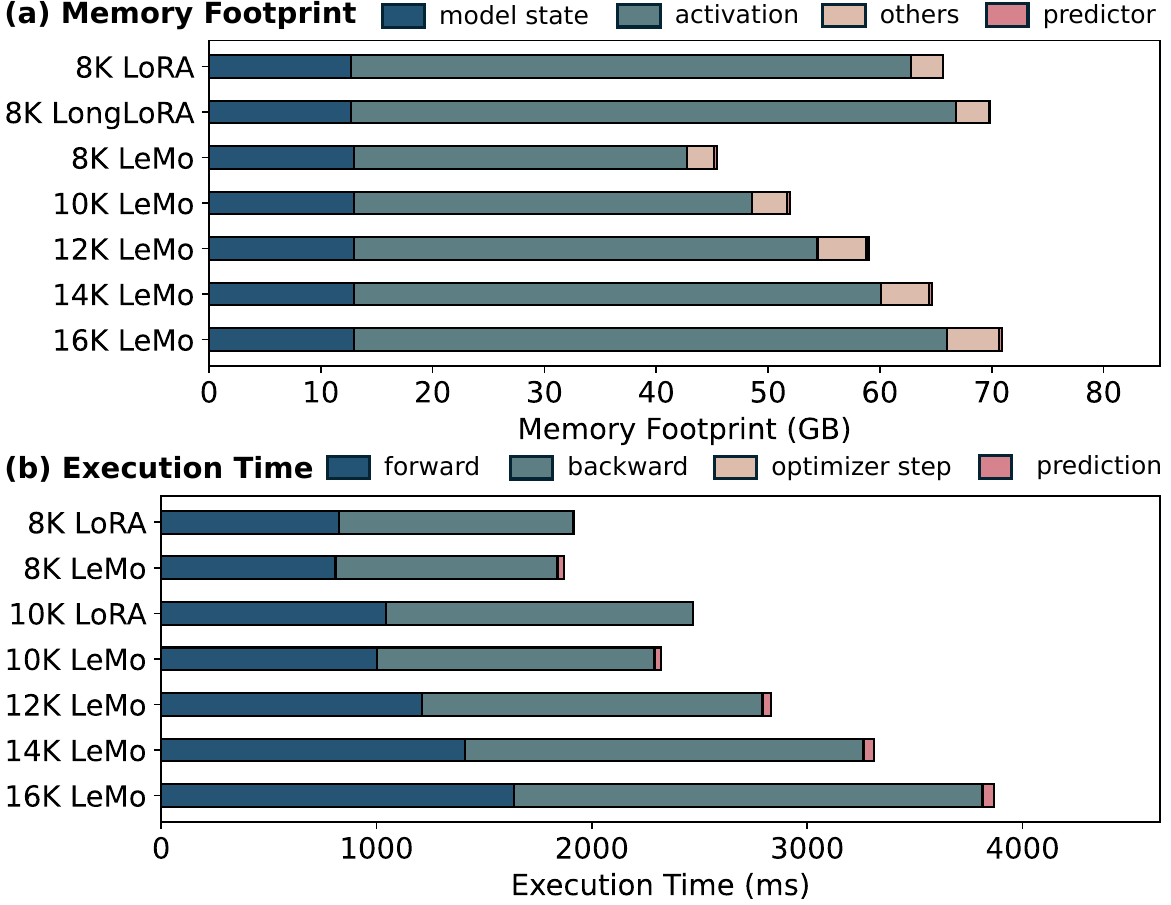}
    \caption{Performance breakdown of Llama2 fine-tuning: (a) Memory footprint and (b) Execution time.}
    \label{fig:exp-ablation-breakdown}
\end{figure}

\noindent\textbf{Fine-grained Performance Breakdown} Figure~\ref{fig:exp-ablation-breakdown} presents a detailed performance breakdown of \LeMo{}, covering both memory and computational aspects. For the memory aspect, the results show that \LeMo{} effectively reduces activation memory consumption compared to both LoRA and LongLoRA. Although the predictors introduced incur additional memory usage, their overhead is minimal, ensuring that the overall memory reduction is preserved. Besides, this reduction is consistent across varying sequence lengths, with the decrease in activation memory scaling linearly with sequence length. For the computational aspect, \LeMo{} also achieves computational gains over LoRA, as the reduced token involvement leads to decreased computation during both the forward and backward phases. Similarly, the computational overhead of predictors is negligible in the context of the overall process.

\begin{figure}
    \centering
    \includegraphics[width=1.0\linewidth]{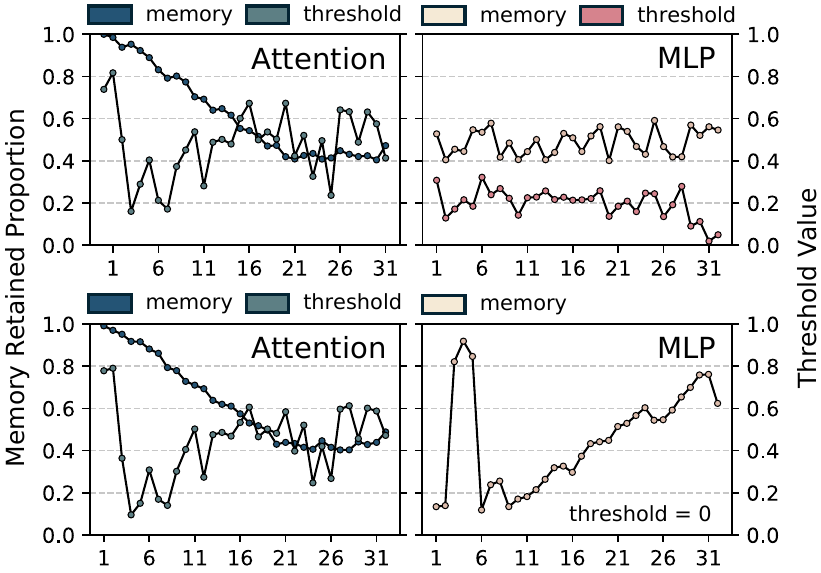}
    \caption{Memory footprint and corresponding threshold across layers on Llama2 7B (upper) and OPT 6.7B (lower).}
    \label{fig:exp-ablation-algorithm}
\end{figure}

\noindent\textbf{Technique 1: Token Elimination.} We delve into the layer level to analyze the effectiveness of our information-driven token elimination algorithm. Figure~\ref{fig:exp-ablation-algorithm} presents the memory consumption and corresponding threshold values across different layers in both self-attention and MLP blocks. Evaluations are conducted on both Llama2 and OPT models, considering their distinct MLP block architecture, which respectively uses SiLU and ReLU as activation functions. The results indicate that the token elimination algorithm achieves average memory savings of 38.3\% (38.0\%) on Attention block and 51.1\% (54.8\%) on MLP block for Llama2 (OPT) model. Besides, the application of layer-specific thresholds allows for varying degrees of reduction across layers, maximizing the exploitation of token-level sparsity while preserving model accuracy.

\begin{figure}
    \centering
    \includegraphics[width=1.0\linewidth]{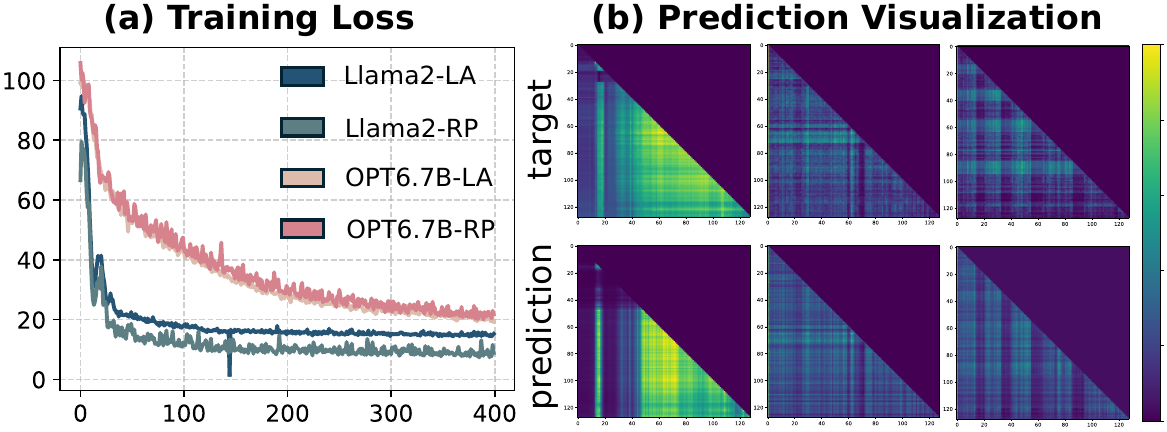}
    \caption{(a) Training loss curve on LongAlign (LA) / RedPajama (RP) and (b) Prediction visualizations of predictors.}
    \label{fig:exp-ablation-predictor}
\end{figure}

\noindent\textbf{Technique 2: Pattern Prediction.} Figure~\ref{fig:exp-ablation-predictor}(a) presents the training loss of predictors across two models and two datasets. The results show that the predictors converge quickly during offline training, requiring fewer than 400 epochs, an acceptable overhead given the following expensive LLM fine-tuning. We then calculate the recall metrics for evaluating the accuracy of predictors, achieving an impressive average of 95.13\%. Particularly, we provide visual comparisons of the predictions against the ground truth. As depicted in Figure~\ref{fig:exp-ablation-predictor}(b), the predicted attention scores closely approximate the ground truth, effectively identifying redundant tokens with high accuracy.

To assess the effectiveness of elastic size transformation, we measure the parameter sizes of predictors across various model layers. Table~\ref{tab:exp-ablation-predictor} details that by exploiting the inherent sparsity within predictors, their parameter sizes can be uniformly reduced across layers, on average 64.6\%. This reduction in predictor size minimizes prediction overhead, aligning consistently with the findings in the performance breakdown.

\begin{table}[b]
    \centering
    \setlength{\tabcolsep}{3pt}
    \caption{Parameter size (in millions) of predictors across layers of Llama2, with or without elastic size transformation.}
    \label{tab:exp-ablation-predictor}
    \small
    \begin{tabular}{llllllll}
        \toprule
        \# Layer & 1      & 5      & 10     & 15     & 20     & 25     & 30     \\ \midrule
        Origin   & 12.58  & 12.58  & 12.58  & 12.58  & 12.58  & 12.58  & 12.58  \\
        Pruned   &  8.19  &  8.19  &  8.13  &  8.10  &  8.08  &  8.07  &  8.10  \\
        Pct.     & 65.1\% & 65.1\% & 64.6\% & 64.5\% & 64.2\% & 64.1\% & 64.5\% \\
        \bottomrule
    \end{tabular}
\end{table}

\begin{figure}
    \centering
    \includegraphics[width=1.0\linewidth]{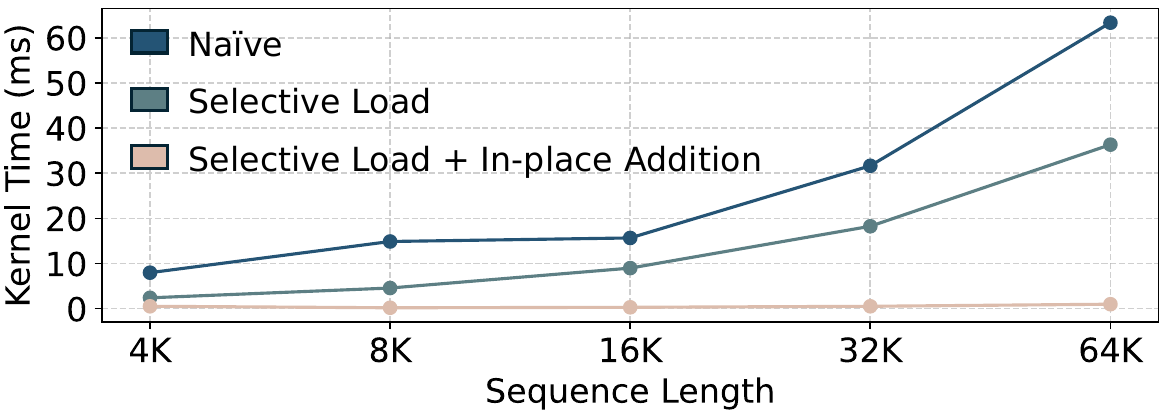}
    \caption{Performance of \LeMo{}'s permutation-free kernel.}
    \label{fig:exp-ablation-kernel-movement}
\end{figure}

\noindent\textbf{Technique 3: Kernel Optimization.} We benchmark the performance of kernels optimized with permutation-free token movement against their naive implementation under various sequence lengths, as depicted in Figure~\ref{fig:exp-ablation-kernel-movement}. The results reveal that both kernel fusion strategies, selective load and in-place addition, effectively enhance performance. The overall speedups increase with sequence length, ranging from $10\times$ to over $50\times$. This improvement primarily stems from the reduction in global memory movement and temporary data allocation. These findings highlight the critical importance of high-performance kernel design, which serves as a robust foundation for the algorithmic framework of \LeMo{}.

\begin{figure}
    \centering
    \includegraphics[width=1.0\linewidth]{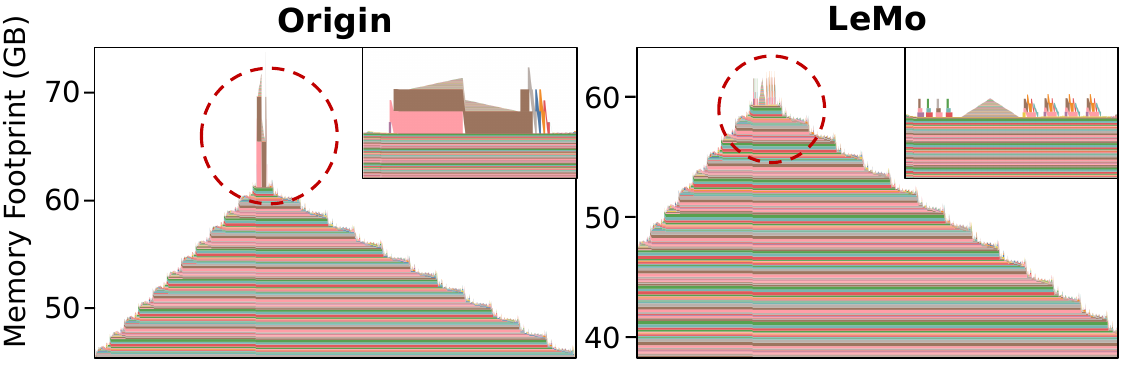}
    \caption{Memory usage peak in loss gradient computation.}
    \label{fig:exp-ablation-kernel-cutting}
\end{figure}

Meanwhile, Figure~\ref{fig:exp-ablation-kernel-cutting} illustrates the memory consumption during fine-tuning with and without the segment-based peak-cutting technique. In the naive implementation, gradient computation requires about 10GB of temporary activation memory, resulting in an inefficient memory peak. By partitioning the gradient computation into smaller segments, the sharp peak is divided into multiple, much smaller memory peaks, achieving an additional 15\% of memory savings.

\subsection{Extension Evaluation}
\begin{figure}
    \centering
    \includegraphics[width=1.0\linewidth]{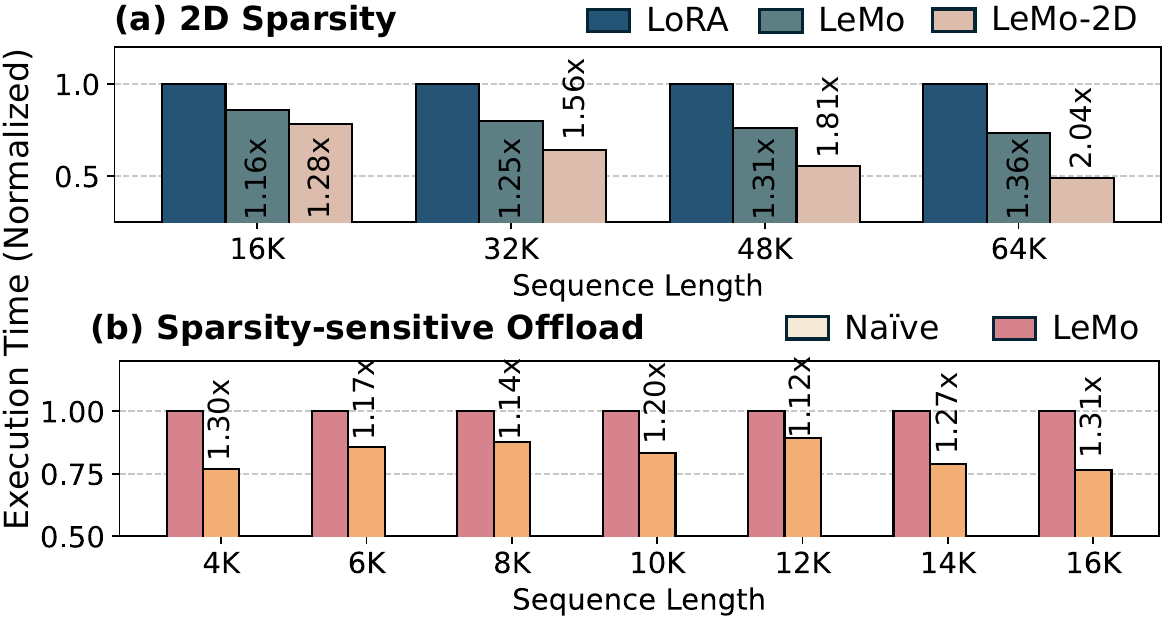}
    \caption{Performance improvements from two extensions.}
    \label{fig:exp-extension}
\end{figure}

\noindent\textbf{Extension 1: Two-dimensional Sparsity.} Building upon \LeMo{}, we explore applying existing hidden-dimension-level sparsity techniques~\cite{long-exposure} to the remaining tokens during attention computation. Figure~\ref{fig:exp-extension}(a) shows that this two-dimensional sparsity further improves the computational efficiency of \LeMo{}, achieving up to $2.04\times$ speedups on Llama2.

\noindent\textbf{Extension 2: Sparsity-sensitive Offload.} We compare the performance of our sparsity-sensitive offloading against a naive uniform offloading strategy. Unlike the naive approach, \LeMo{} considers the varying sparsity ratios across layers, allowing for the offloading of more activations or a reduction in data transfer latency. Figure~\ref{fig:exp-extension}(b) shows that this technique achieves an average speedup of $1.22\times$ speedups on Llama2.

\subsection{Scalability Analysis}
\begin{figure}
    \centering
    \includegraphics[width=1.0\linewidth]{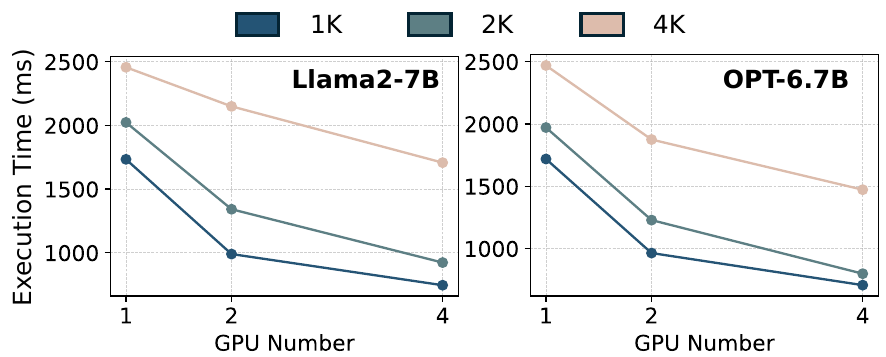}
    \caption{Strong scalability evaluation of \LeMo{}.}
    \label{fig:exp-scalability}
\end{figure}

We conclude by analyzing the strong scalability of \LeMo{} on 4$\times$4090 GPUs. Figure~\ref{fig:exp-scalability} shows that \LeMo{}'s performance scales proportionally with GPU number across different models and sequence lengths. The scalability is achieved because \LeMo{} seamlessly minimizes token involvements and introduces no extra communication overhead. These results highlight \LeMo{}'s potential for deployment in large-scale systems.

\section{Related Work}

\noindent\textbf{Optimizations for Activation Memory.} As a primary memory bottleneck in LLM training or fine-tuning, activation memory consumption has been the focus of extensive research, which can be categorized into three main approaches: The first is \textit{activation recomputation}~\cite{activation-recompute,activation-recompute-2,activation-recompute-3}, designed to avoid storing activations during the forward pass but recomputing them during the backward pass. The second is \textit{activation offloading}~\cite{autotm,swapadvisor,capuchin,l2l}, which asynchronously transfers activations from GPU to CPU and prefetches them back before required. The last is \textit{activation compression}~\cite{actnn,ac-gc,gact}, which reduces the activation memory size through quantization or pruning. However, these methods primarily trade memory for additional computation or communication, rather than fundamentally reducing memory demands. In contrast, \LeMo{} directly minimizes activation memory requirements and can be seamlessly combined with all these optimizations.

\noindent\textbf{Optimizations for Long-context Fine-tuning.} To effectively extend the context window to longer sequences, some methods~\cite{position-interpolation,longrope,scaled-rope} focus on optimizing the fine-tuning algorithm design. Besides, some methods~\cite{code-llama,llama-long,focused-transformer} explore strategies for modifying the position embeddings of LLMs to handle longer context. All these efforts are complementary to \LeMo{}.

Beyond effectiveness, some recent methods~\cite{pose,activation-beacon} are proposed to mitigate the substantial fine-tuning overheads for efficiency. Particularly, Parameter-efficient fine-tuning methods~\cite{lora,adapter,bitfit,prefix-tuning} first offer an effective solution by reducing the number of trainable parameters and memory usage while achieving comparable performance to full fine-tuning. Furthermore, a series of sparsity-based methods~\cite{longlora,long-exposure} are proposed to further reduce the computation costs by exploiting the inherent sparsity within attention mechanism~\cite{sparse-attention}. However, while PEFT methods greatly cut down the memory consumption of optimizer states, the activation memory emerges as the primary bottleneck. Although existing sparsity mechanisms deliver notable computational gains, the presence of shadowy activation prevents comparable benefits for activation memory. Instead, \LeMo{} achieves the best of both worlds by identifying and exploiting contextual token sparsity.

\noindent\textbf{Optimizations for Token Utilization.} Sharing the same high-level idea of \LeMo{}, several studies also explore leveraging the inherent redundancy that existed in natural language, including data engineering~\cite{data-engineering,metadata}, prompt compression~\cite{selective-context,llmlingua}, and inference optimization~\cite{flexgen,infinigen,h2o,quest}. Notably, some works~\cite{length-adaptive-transformer,tr-bert,adapler,learned-token-pruning,dynamicvit,token-merging,powerbert} propose eliminating tokens during inference to reduce model latency. However, these methods are designed for smaller models and all focus solely on model inference. Instead, \LeMo{} is the first to optimize LLM long-context fine-tuning to our best knowledge.

\section{Conclusion}
We propose \LeMo{}, an efficient system designed to optimize long-context fine-tuning for LLMs. Our approach introduces a novel sparsity mechanism within LLM long-context fine-tuning, termed contextual token sparsity. To systematically exploit this mechanism, we develop three key techniques that identify, predict, and exploit this sparsity, achieving both memory savings and performance speedups over state-of-the-art methods. Compression embodies intelligence, with sparsity serving as a potent form of compression. We envision \LeMo{} inspiring broader exploration of sparsity for advancing LLMs.

\bibliographystyle{plain}
\bibliography{references}

\end{document}